\documentclass{article}

\PassOptionsToPackage{numbers, comma, compress, sort}{natbib}

\usepackage[preprint]{neurips_2024}

\usepackage[utf8]{inputenc} 
\usepackage[T1]{fontenc}    
\usepackage{hyperref}       
\usepackage{url}            
\usepackage{booktabs}       
\usepackage{amsfonts}       
\usepackage{nicefrac}       
\usepackage{microtype}      
\usepackage{xcolor}         

\usepackage{amsmath}
\usepackage{amssymb}
\usepackage{amsthm}

 \usepackage{graphicx} 
\usepackage{arydshln}
\usepackage{natbib}

\title{Rapid Adaptation of Earth Observation Foundation Models for Segmentation}

\author{%
  Karthick Panner Selvam\\
  University of Luxembourg\\
  \texttt{karthick.pannerselvam@uni.lu} 
  \And
  Raúl Ramos-Pollán \\
  Universidad de Antioquia, Colombia\\
  \texttt{raul.ramos@udea.edu.co}
  \And
  Freddie Kalaitzis \\
  University of Oxford, UK\\
  \texttt{freddie.kalaitzis@cs.ox.ac.uk} 
  }

\begin{document}

\maketitle

\begin{abstract}
This study investigates the efficacy of Low-Rank Adaptation (LoRA) in fine-tuning Earth Observation (EO) foundation models for flood segmentation. We hypothesize that LoRA, a parameter-efficient technique, can significantly accelerate the adaptation of large-scale EO models to this critical task while maintaining high performance.
We apply LoRA to fine-tune a state-of-the-art EO foundation model pre-trained on diverse satellite imagery, using a curated dataset of flood events. Our results demonstrate that LoRA-based fine-tuning (r-256) improves F1 score by 6.66 points and IoU by 0.11 compared to a frozen encoder baseline, while significantly reducing computational costs. Notably, LoRA outperforms full fine-tuning, which proves computationally infeasible on our hardware.
We further assess generalization through out-of-distribution (OOD) testing on a geographically distinct flood event. While LoRA configurations show improved OOD performance over the baseline.
This work contributes to research on efficient adaptation of foundation models for specialized EO tasks, with implications for rapid response systems in disaster management. Our findings demonstrate LoRA's potential for enabling faster deployment of accurate flood segmentation models in resource-constrained, time-critical scenarios.
\end{abstract}

\section{Introduction}
The advent of large-scale foundation models in Earth Observation (EO) has revolutionized our capacity to extract meaningful insights from satellite imagery\cite{smith2023earthpt,guo2024skysense}. These models, often based on transformer architectures and trained on vast, diverse datasets, have demonstrated remarkable zero-shot and few-shot learning capabilities across a spectrum of remote sensing tasks\cite{vaswani2023attentionneed}\cite{allen2023fewshotlearningglobalmultimodal}\cite{zheng2024segmentchange}. However, the adaptation of such models to specific, high-stakes applications like flood segmentation presents unique challenges, particularly in terms of computational efficiency and rapid deployment.

Flooding events, exacerbated by climate change, pose an increasing threat to global communities and ecosystems. The ability to quickly and accurately delineate flood extents from satellite imagery is crucial for effective disaster response and mitigation strategies \cite{portales-julia_global_2023}\cite{reed2022impact}\cite{konapala2021exploring}\cite{10020911}. Traditional approaches to flood detection, including thresholding techniques on Synthetic Aperture Radar (SAR) data or machine learning models trained on optical imagery, often struggle with the temporal and spatial heterogeneity inherent in flood events.

Foundation models in EO \cite{jakubik2023foundationmodelsgeneralistgeospatial} \cite{cong2022satmae}  \cite{stewart2023ssl4eo} \cite{guo2024skysense} \cite{smith2024earthpttimeseriesfoundation}  \cite{allen2023largescalemaskedautoencoding} \cite{allen2023fewshotlearningglobalmultimodal} \cite{10490262}, have shown promise in addressing these challenges through their ability to leverage pre-trained representations for downstream tasks. However, the conventional fine-tuning of these models for specific applications like flood segmentation is computationally intensive and time-consuming, often requiring substantial GPU resources and large amounts of labeled data.

Low-Rank Adaptation (LoRA), introduced by \cite{hu2021loralowrankadaptationlarge}, offers a potential solution to these limitations. LoRA operates by introducing trainable low-rank decomposition matrices into the model, allowing for task-specific adaptation with minimal additional parameters. This approach has shown remarkable success in Natural Language Processing (NLP) domains, enabling efficient fine-tuning of large language models \cite{mao2024surveyloralargelanguage}. However, its application to EO foundation models, particularly for dense prediction tasks like segmentation, remains largely unexplored.

Our research investigates the efficacy of LoRA in adapting EO foundation models for flood segmentation. We hypothesize that LoRA can significantly reduce the computational overhead of fine-tuning while maintaining or even improving performance compared to traditional methods. To test this hypothesis, we employ a state-of-the-art EO foundation model pre-trained on a diverse corpus of satellite imagery, including multi-spectral, SAR.
We implement LoRA by introducing low-rank update matrices in the attention layers of our transformer-based foundation model. The rank of these matrices is treated as a hyperparameter, allowing us to explore the trade-off between model capacity and computational efficiency.

We evaluate model performance using metrics standard in the field of semantic segmentation, including Intersection over Union (IoU) and F1 score. To quantify the efficiency gains of LoRA, we meticulously measure GPU memory usage, and the number of trainable parameters. We also analyze the impact of different LoRA configurations on model performance, investigating how the rank of update matrices and their placement within the model architecture affect segmentation accuracy and training dynamics. This analysis provides insights into the key factors driving the performance of LoRA-adapted models in the context of flood detection.

\section{Background}

\subsection{Earth Observation Foundation Models}

Earth Observation (EO) foundation models represent a paradigm shift in remote sensing, leveraging large-scale pre-training on diverse satellite imagery to learn generalizable representations. These models, often based on transformer architectures, can be formalized as follows:

Let $\mathcal{X} = \{x_1, ..., x_N\}$ be a set of input satellite image patches, where each $x_i \in \mathbb{R}^{C \times H \times W}$ represents a patch with $C$ channels and spatial dimensions $H \times W$. A transformer-based EO foundation model $\mathcal{F}$ can be described as a function:

\begin{equation}
    \mathcal{F}: \mathcal{X} \rightarrow \mathcal{Z}
\end{equation}

where $\mathcal{Z} = \{z_1, ..., z_N\}$ and $z_i \in \mathbb{R}^{D}$ is the learned representation for patch $x_i$.

The core of the transformer architecture in these models is the self-attention mechanism \cite{vaswani2023attentionneed}. For a given query $Q$, key $K$, and value $V$, the self-attention operation is defined as:

\begin{equation}
    \text{Attention}(Q, K, V) = \text{softmax}\left(\frac{QK^T}{\sqrt{d_k}}\right)V
\end{equation}

where $d_k$ is the dimension of the key vectors.

\begin{figure}[t]
\centering
\includegraphics[width=\textwidth]{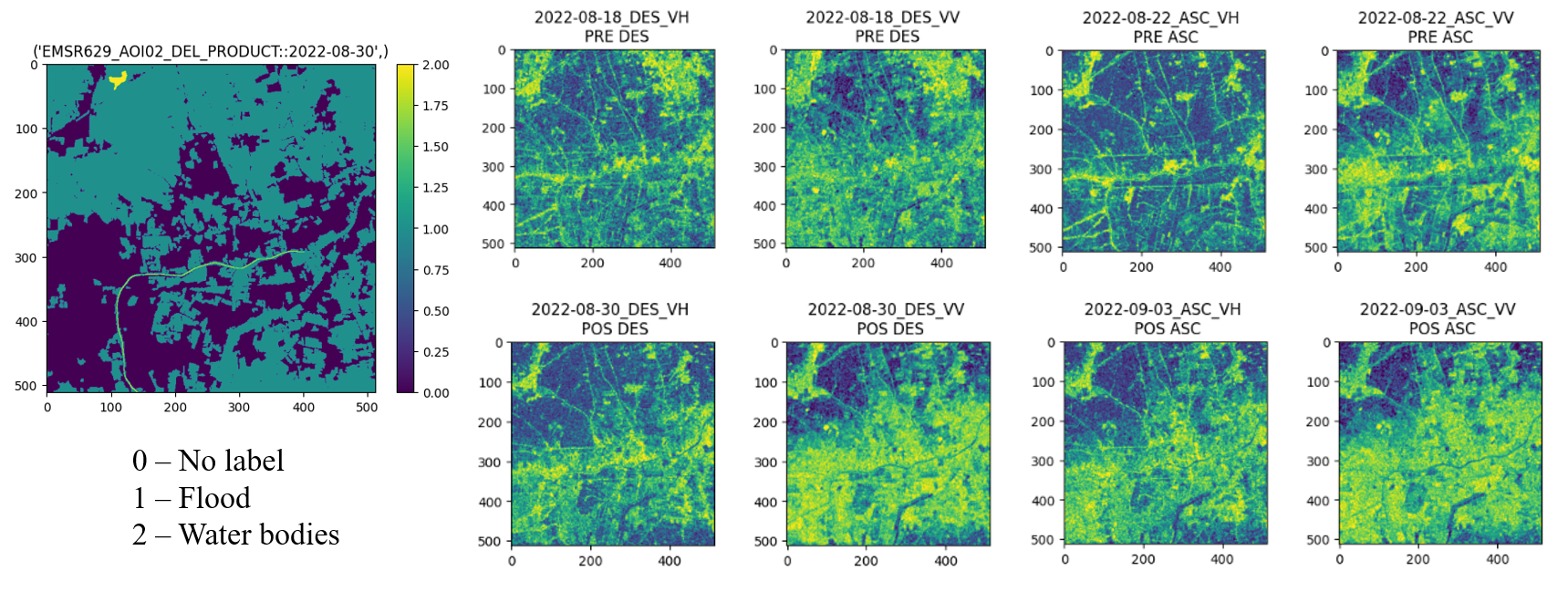}
\caption{Dataset composition for flood detection. Each sample includes pre-event and post-event Sentinel-1 SAR imagery (VV and VH polarizations for both ascending and descending orbits) along with corresponding flood extent labels. Pre: pre-event; Post: post-event; VV: vertical transmit-vertical receive polarization; VH: vertical transmit-horizontal receive polarization.}\label{fig:dataset}
\end{figure}

\subsection{Theoretical Analysis of Low-Rank Adaptation (LoRA)}

LoRA is a parameter-efficient fine-tuning method that addresses the challenge of adapting large pre-trained models to specific tasks or domains. LoRA introduces trainable low-rank decomposition matrices into pre-trained models, significantly reducing the number of trainable parameters while maintaining model performance. For a pre-trained weight matrix $W \in \mathbb{R}^{d \times k}$, LoRA defines the update as:

\begin{equation}
    W' = W + BA
\end{equation}

where $B \in \mathbb{R}^{d \times r}$ and $A \in \mathbb{R}^{r \times k}$ are low-rank matrices, and $r \ll \min(d, k)$ is the rank of the update.

In practice, LoRA initializes $B$ and $A$ randomly and learns them through gradient descent, as empirically validated by \cite{hu2021loralowrankadaptationlarge}. The optimization objective for LoRA can be formulated as:

\begin{equation}
    \min_{B, A} \mathcal{L}(\mathcal{F}(\mathcal{X}; \theta, BA), \mathcal{Y})
\end{equation}

where $\mathcal{L}$ is the task-specific loss function, $\mathcal{F}$ is the model, $\theta$ are the frozen pre-trained parameters, $\mathcal{X}$ is the input data, and $\mathcal{Y}$ are the target labels. The gradient updates for the LoRA parameters follow:

\begin{align}
    B &\leftarrow B - \eta \frac{\partial \mathcal{L}}{\partial B} \\
    A &\leftarrow A - \eta \frac{\partial \mathcal{L}}{\partial A}
\end{align}

where $\eta$ is the learning rate. LoRA's parameter efficiency becomes evident when we consider the number of trainable parameters. For a weight matrix $W \in \mathbb{R}^{d \times k}$, full fine-tuning would require updating all $dk$ parameters. In contrast, LoRA only updates $r(d + k)$ parameters, which is significantly smaller when $r \ll \min(d, k)$. This reduction in trainable parameters has been shown to lead to faster convergence and improved generalization in various tasks \cite{hu2021loralowrankadaptationlarge}.

The computational advantage of LoRA is particularly pronounced in transformer architectures \cite{vaswani2023attentionneed}, where it is typically applied to the query, key, and value projection matrices in the self-attention mechanism. For a transformer layer with embedding dimension $d_{\text{model}}$ and key dimension $d_k$, the LoRA updates are:

\begin{align}
    W_q' &= W_q + B_qA_q \\
    W_k' &= W_k + B_kA_k \\
    W_v' &= W_v + B_vA_v
\end{align}

where $W_q, W_k, W_v \in \mathbb{R}^{d_{\text{model}} \times d_k}$ are the original projection matrices, and $B_q, B_k, B_v \in \mathbb{R}^{d_{\text{model}} \times r}$ and $A_q, A_k, A_v \in \mathbb{R}^{r \times d_k}$ are the LoRA update matrices.

\begin{figure}[t]
\centering
\includegraphics[width=\textwidth]{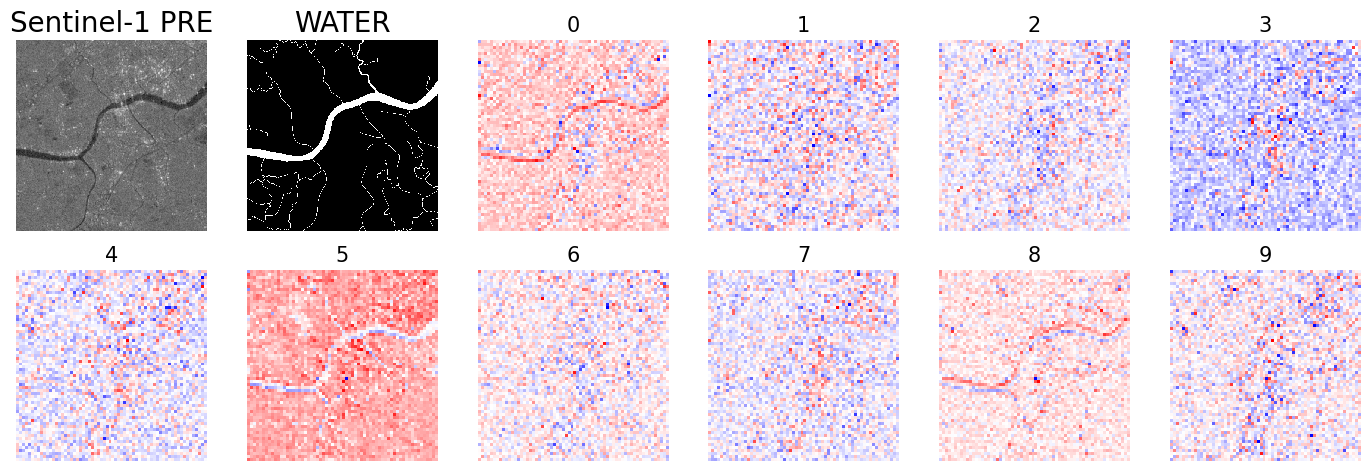}
\caption{Visualization of patch embeddings from the pre-trained Clay EO-FM encoder using t-SNE projection. The projections consistently demonstrate clear clustering of water and land regions, indicating the encoder's ability to distinguish these surface types in the latent space. } 
\label{fig:emb}
\end{figure}

\section{Methodology}

We propose an efficient approach to adapt the Clay\footnote{https://github.com/Clay-foundation/model} EO-FM for flood detection using LoRA. Our method leverages Sentinel-1 SAR data to ensure robust performance across all weather conditions.

\subsection{Problem Formulation}

Let $\mathcal{X} = \{X_1, \ldots, X_N\}$ be a set of input satellite image pairs, where each $X_i = (X_i^{\text{pre}}, X_i^{\text{post}}) \in \mathbb{R}^{2 \times C \times H \times W}$ represents pre-flood and post-flood images with $C=4$ channels (Sentinel-1 VV and VH for ascending and descending orbits) and spatial dimensions $H = W = 512$ at sentinels resolution of 10m/pixels, corresponds to a spatial resolution of 5.12km$\times$5.12km, or 26.2km$^2$ as shown in Figure \ref{fig:dataset}. The corresponding flood masks are denoted as $\mathcal{Y} = \{Y_1, \ldots, Y_N\}$, where $Y_i \in \{0, 1\}^{H \times W}$. Our objective is to learn a function $f_\theta: \mathcal{X} \rightarrow \mathcal{Y}$ that accurately maps input image pairs to flood extent masks. Formally, we aim to solve:

\begin{equation}
    \theta^* = \min_\theta \mathbb{E}_{(X,Y) \sim \mathcal{D}} [\mathcal{L}(f_\theta(X), Y)]
\end{equation}

where $\mathcal{D}$ is the data distribution and $\mathcal{L}$ is a suitable loss function.

\subsection{Model Architecture}

We leverage the encoder component of the Clay EO-FM, which is based on the Vision Transformer (ViT) architecture \cite{dosovitskiy2021image} and trained using masked autoencoder objectives \cite{he_masked_2021}. This pre-trained encoder is combined with our custom deconvolution network for the segmentation task. The complete model can be formulated as:

\begin{align}
    Z &= E_{\text{Clay}}(X; \theta_E) \\
    \hat{Y} &= D_{\text{custom}}(Z; \theta_D)
\end{align}

where $E_{\text{Clay}}$ is the pre-trained Clay EO-FM encoder, $Z$ represents the encoded features, $D_{\text{custom}}$ is our custom deconvolution network, and $\hat{Y}$ is the predicted flood mask. We conducted a qualitative analysis of the latent space representations generated by the pre-trained encoder for the input SAR data. Figure \ref{fig:emb} illustrates the t-SNE \cite{JMLR:v9:vandermaaten08a} projection of these high-dimensional patch embeddings. The visualization reveals distinct clustering patterns that correspond to water and land regions, indicating that the encoder effectively captures and differentiates the intrinsic characteristics of these surface types in the embedding space. This spatial coherence in the latent representations suggests that the pre-trained model has learned meaningful features relevant to flood detection tasks, even before fine-tuning.

The Clay EO-FM encoder consists of 12 transformer layers with 12 attention heads each, and a hidden dimension $d_{\text{model}} = 768$. Our deconvolution network $D_{\text{custom}}$ consists of two convolutional layers followed by three deconvolutional layers, each paired with ReLU activation \cite{ronneberger2015unetconvolutionalnetworksbiomedical} \cite{agarap2019deeplearningusingrectified}. This structure progressively upsamples the encoded features from 64x64 to 512x512 resolution, enabling detailed flood extent segmentation. The outputs for all input directions and seasons are concatenated and passed through a final convolutional layer to produce the segmentation logits. We apply LoRA to the attention mechanism in each transformer layer of the Clay EO-FM encoder. Specifically, we adapt the $\text{to\_qkv}$ and $\text{to\_out}$ linear transformations in the Attention module. For a pre-trained weight matrix $W \in \mathbb{R}^{d_{\text{in}} \times d_{\text{out}}}$, LoRA defines the update:

\begin{equation}
    W' = W + \alpha \frac{BA}{\text{rank}(BA)}
\end{equation}

where $B \in \mathbb{R}^{d_{\text{in}} \times r}$, $A \in \mathbb{R}^{r \times d_{\text{out}}}$, $\alpha$ is a scaling factor, and $r$ is the LoRA rank. In our implementation, the attention mechanism is defined as:

\begin{align}
    \text{QKV} &= \text{to\_qkv}(X) \\
    Q, K, V &= \text{chunk}(\text{QKV}, 3) \\
    \text{Attention}(Q, K, V) &= \text{to\_out}(\text{softmax}(\frac{QK^T}{\sqrt{d_{\text{head}}}})V)
\end{align}

where $X \in \mathbb{R}^{b \times (h \times w) \times d_{\text{model}}}$ is the input to the attention layer, $b$ is the batch size, $h \times w$ is the number of patches (which effectively becomes the sequence length), and $d_{\text{model}}$ is the model dimension. For an input image of size $H \times W$ and patch size $P \times P$, we have $h = H/P$ and $w = W/P$. LoRA is applied to both $\text{to\_qkv}$ and $\text{to\_out}$ transformations:

\begin{align}
    \text{to\_qkv}'(X) &= (W_{\text{qkv}} + \alpha \frac{B_{\text{qkv}}A_{\text{qkv}}}{\text{rank}(B_{\text{qkv}}A_{\text{qkv}})})X \\
    \text{to\_out}'(X) &= (W_{\text{out}} + \alpha \frac{B_{\text{out}}A_{\text{out}}}{\text{rank}(B_{\text{out}}A_{\text{out}})})X
\end{align}

This approach allows us to efficiently adapt the attention mechanism while keeping most of the pre-trained weights frozen.

\subsection{Training Objective}

We optimize the model using a combination of binary cross-entropy (BCE) and Dice loss:

\begin{equation}
    \mathcal{L} = \lambda_{\text{BCE}} \mathcal{L}_{\text{BCE}} + \lambda_{\text{Dice}} \mathcal{L}_{\text{Dice}}
\end{equation}

where:

\begin{align}
    \mathcal{L}_{\text{BCE}} &= -\frac{1}{N}\sum_{i=1}^N [y_i \log(\hat{y}_i) + (1-y_i) \log(1-\hat{y}_i)] \\
    \mathcal{L}_{\text{Dice}} &= 1 - \frac{2\sum_{i=1}^N y_i\hat{y}_i + \epsilon}{\sum_{i=1}^N y_i^2 + \sum_{i=1}^N \hat{y}_i^2 + \epsilon}
\end{align}

with $\lambda_{\text{BCE}} = \lambda_{\text{Dice}} = 1$ as weighting factors, and $\epsilon$ as a small constant for numerical stability.

This methodology enables efficient adaptation of the Clay EO-FM encoder for flood detection using Sentinel-1 SAR data, leveraging the parameter-efficient LoRA technique while maintaining the model's pre-trained knowledge. The custom deconvolution network then transforms the encoded features into the final segmentation mask.

\section{Experiments and Results}

We conducted experiments to evaluate the effectiveness of our LoRA-based approach in adapting the Clay EO-FM encoder for flood detection. All experiments were performed on an NVIDIA A100 GPU with 40GB memory. We utilized the WorldFlood dataset \cite{portales-julia_global_2023}, originally created for Sentinel-2 imagery, which we modified to use Sentinel-1 SAR data for more accurate flood detection under all weather conditions. We maintained the original train, validation, and test splits from WorldFlood. Additionally, we created an out-of-distribution (OOD) test set using data from the Luxembourg flood event of July 2021 to evaluate our model's generalization ability.

We compared three adaptation strategies: (1) Full fine-tuning of the Clay encoder and our custom decoder, (2) Freezing the encoder and training only the decoder, (3) LoRA fine-tuning, updating LoRA parameters and the decoder. We used this setting for all models, utilizing a maximum of 50 epochs with an early stopping mechanism triggered after 2 epochs of stagnant validation loss. The optimization process was governed by the Adam optimizer with an initial learning rate of $10^{-4}$ and weight decay of $10^{-4}$. To adapt to the learning dynamics, we implemented a ReduceLROnPlateau scheduler with a decay factor of 0.5 and a patience of 3 epochs. For the LoRA configurations, we set the $\alpha$ parameter to twice the rank ($\alpha = 2r$) and applied a dropout rate of 0.1 to the LoRA layers.

For the LoRA fine-tuning strategy, we explored different rank values ($r$) to investigate the trade-off between model performance and computational efficiency. Table~\ref{tab:results} presents a comprehensive comparison of these strategies across various metrics.

\begin{table}[h]
\centering
\caption{Performance comparison of adaptation strategies for flood detection using Clay EO-FM on the in-distribution test set.}

\begin{tabular}{lcccccc}
\hline
\hline
Method                  & F1             & Accuracy       & Precision      & IoU           & GPU Mem. & Train. params \\ \hline \hline
Full Fine-tuning        & -              & -              & -              & -             & OOM        & 123.94M                 \\
Frozen Encoder          & 84.20          & 89.37          & 74.32          & 0.72          & 17.46 GB   & 8.65M                  \\
LoRA Fine-tuning r-8    & 80.54          & 86.11          & 68.11          & 0.67          & 17.54 GB           & 9.22M            \\
LoRA Fine-tuning r-64   & 84.16          & 89.19          & 73.49          & 0.72          & 18.96 GB            & 13.18M           \\
LoRA Fine-tuning r-256  & \textbf{90.86} & \textbf{94.29} & 85.31          & \textbf{0.83} & 19.45 GB   & 26.74M           \\
LoRA Fine-tuning r-512  & 89.10          & 93.14          & 83.07          & 0.80          & 19.05 GB           & 44.83M           \\
LoRA Fine-tuning r-1024 & 89.87          & 93.88          & \textbf{86.95} & 0.81          & 24.39 GB   & 81.01M           \\ \hline
\end{tabular}
\label{tab:results}
\end{table}

Our experiments reveal several key findings:  Attempts to fully fine-tune the Clay encoder and custom decoder resulted in out-of-memory (OOM) errors, even with a 40GB GPU. This highlights the challenge of adapting large-scale models with limited computational resources.  Freezing the encoder and training only the decoder provided a strong baseline, achieving an F1 score of 84.20 and IoU of 0.72 with minimal GPU memory usage (17.46 GB) and trainable parameters (8.65M). LoRA fine-tuning demonstrated superior performance across various rank values. Notably, LoRA with rank 256 achieved the best overall performance, with an F1 score of 90.86, accuracy of 94.29\%, and IoU of 0.83, surpassing both the frozen encoder baseline and other LoRA configurations.
    
As the LoRA rank increased, we observed a general trend of improved performance at the cost of increased GPU memory usage and trainable parameters. However, this relationship was not strictly linear, with rank 256 providing the optimal balance. All LoRA configurations used significantly fewer trainable parameters compared to full fine-tuning (123.94M), with the best-performing LoRA r-256 using only 26.74M parameters.

These results demonstrate the effectiveness of our LoRA-based approach in adapting the Clay EO-FM for flood detection, achieving superior performance while maintaining computational efficiency.

\subsection{Out-of-Distribution Generalization}

To assess our model's generalization capability, we evaluated its performance on an OOD test set from the Luxembourg flood event of July 2021, a region not represented in the training data. This evaluation revealed crucial insights into the model's adaptability to new geographical contexts.

As shown in Table \ref{tab:ood-results}, the OOD performance on the Luxembourg flood event dataset reveals interesting insights. Initially, the high accuracy (>98\%) across all configurations appeared promising. However, this metric proved misleading due to the predominance of non-flood pixels, a common challenge in flood detection tasks. The F1 scores and IoU metrics, ranging from 36.50 to 48.60 and 22.33 to 32.10 respectively, provided a more realistic assessment of performance, showing a notable drop compared to in-distribution results.

Interestingly, we observed a precision-recall trade-off as LoRA rank increased. Higher ranks demonstrated improved precision at the cost of recall, suggesting a more conservative flood detection approach in unfamiliar territories. LoRA r-1024 achieved the best overall OOD performance with an F1 score of 48.60 and IoU of 32.10, despite not being the top performer on the in-distribution set. This suggests that higher-rank LoRA adaptations might capture more generalizable features, albeit at the cost of increased computational resources.

\begin{figure}[t]
\centering
\includegraphics[width=0.8\textwidth]{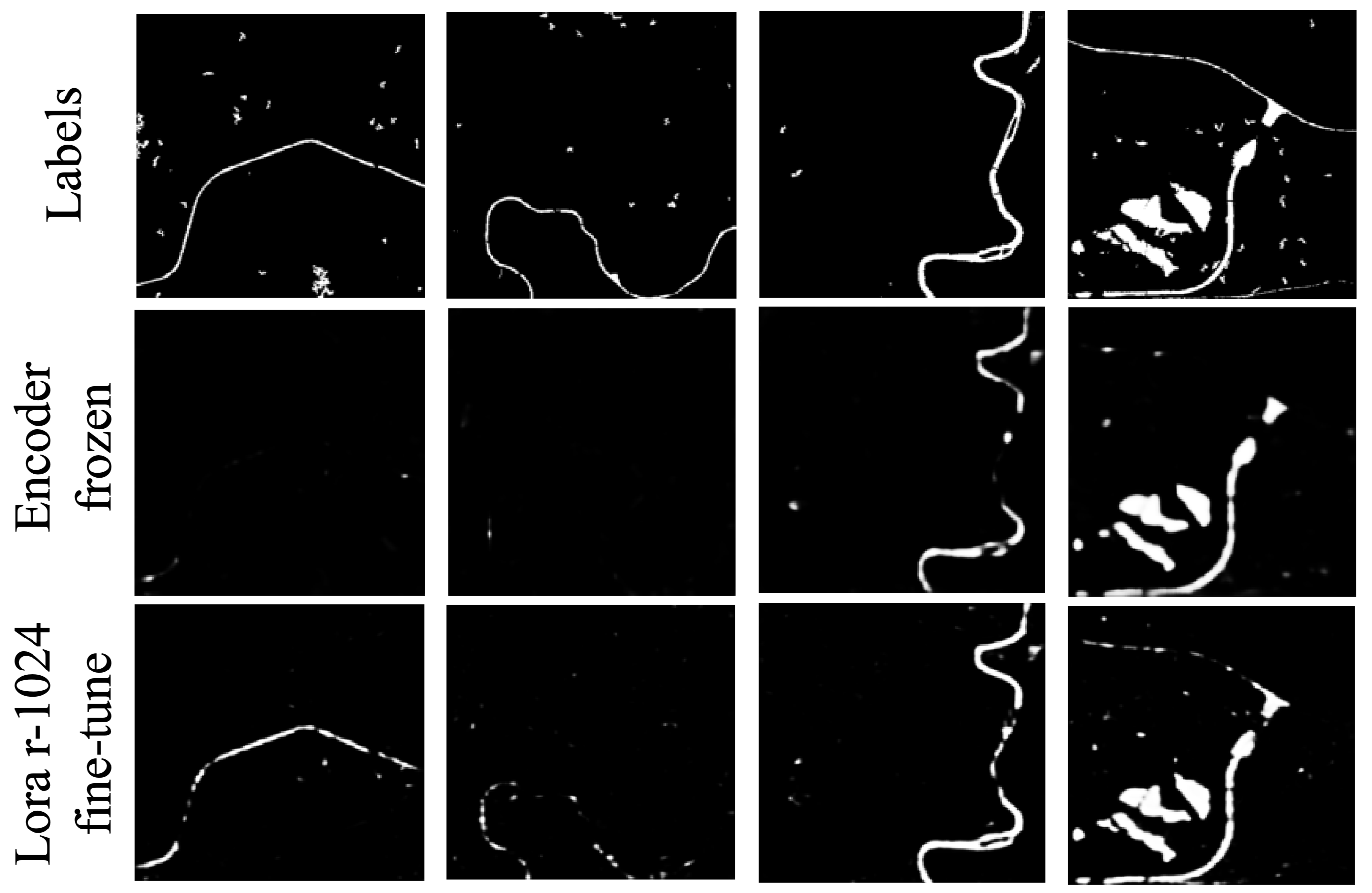}
\caption{Comparison of ground truth and model-predicted flood segmentation on the out-of-distribution Luxembourg flood event dataset. The visualization demonstrates that the LoRA approach generalizes better than the encoder-frozen approach to unseen geographical regions and flood events.}
\label{ood-viz}
\end{figure}

The performance gap between in-distribution and OOD results underscores the challenges in developing flood detection models that can seamlessly adapt to diverse geographical contexts. However, the fact that LoRA configurations generally outperformed the frozen encoder baseline on OOD data indicates that our approach enhances the model's generalization capabilities to some extent as shown in Figure \ref{ood-viz}.

\begin{table}[t]
\caption{Out-of-distribution performance on Luxembourg flood event data for different LoRA configurations and frozen encoder baseline.}
\label{tab:ood-results}
\centering
\begin{tabular}{lcccccc}
\hline \hline
Method & Accuracy & Precision & Recall & F1 & IoU & Dice \\
\hline \hline
Frozen Encoder & 99.18 & 82.29 & 27.38 & 41.09 & 25.86 & 41.09 \\
LoRA r-8 & 98.70 & 37.00 & 36.02 & 36.50 & 22.33 & 36.50 \\
LoRA r-64 & 99.13 & 65.42 & 34.64 & 45.30 & 29.28 & 45.30 \\
LoRA r-256 & 99.21 & \textbf{81.31} & 31.59 & 45.50 & 29.45 & 45.50 \\
LoRA r-512 & 99.19 & 86.26 & 26.42 & 40.45 & 25.35 & 40.45 \\
LoRA r-1024 & \textbf{99.20} & 73.08 & \textbf{36.40} & \textbf{48.60} & \textbf{32.10} & \textbf{48.60} \\
\hline
\end{tabular}
\end{table}

\section{Conclusion}
We presented an efficient approach for adapting a large-scale EO-FM for flood detection using LoRA. Our method demonstrated superior performance on in-distribution data compared to both full fine-tuning and frozen encoder baselines, with LoRA (r-256) improving F1 score by 6.66 points and IoU by 0.11. This approach offers a favorable trade-off between performance and computational efficiency, crucial for resource-constrained scenarios in Earth observation.

Our OOD experiments on the Luxembourg flood event revealed the challenges of generalization in flood detection tasks. While higher-rank LoRA adaptations showed improved OOD performance. These findings underscore the potential of parameter-efficient fine-tuning techniques in making large-scale AI models more accessible and adaptable for critical EO tasks. They also emphasize the need for further research into enhancing model robustness across varied geographical regions and flood event characteristics. Future work could explore applications to diverse EO tasks, investigate methods to improve OOD generalization, and further optimize LoRA configurations for enhanced efficiency and adaptability in environmental monitoring systems.

\begin{ack}
This work has been enabled by FDL Europe | Earth Systems Lab (https://fdleurope.org) a public / private partnership between the European Space Agency (ESA), Luxembourg Space Agency, Trillium Technologies, the University of Oxford in partnership with Google Cloud, NVIDIA Corporation, RSS Hydro, LuxProvide. We are thankful to the SAR-FM FDL 2024 Team and all the reviewers that participated in it.

\end{ack}

\bibliographystyle{unsrtnat}  
\bibliography{ref} 

\end{document}